\pdfoutput=1
\documentclass[11pt,a4paper]{article}
\usepackage[margin=25mm,top=25mm,headheight=41pt,headsep=8pt]{geometry}
\usepackage{styles/lab-preprint}
\usepackage{times}
\usepackage{multirow,bm,booktabs,array,natbib,url}
\newcommand{\PaperTitle}{PartiGen: Part-Aware Representations for Language-Conditioned Humanoid Motion Generation}
\newcommand{\PaperAuthorNames}{Huang Chuxuan, Zhanxiang Cao, Tianyue Zhan, Haoyang Li, Yang Zhang, Liyun Yan, Chenlin Ming, Keqi Zhu, Yizhi Chen, Shengcheng Fu, Zihao Zhou, Xinyi Peng, ZHANG Kaixuan, Boyang Xing, Jiawei Bao, Yang Yang, Yue Gao}
\newcommand{\PaperAuthors}{%
Huang Chuxuan\textsuperscript{1,2,\textdagger}, Zhanxiang Cao\textsuperscript{3,2,\textdagger}, Tianyue Zhan\textsuperscript{3,2}, Haoyang Li\textsuperscript{3,2}, Yang Zhang\textsuperscript{3}, Liyun Yan\textsuperscript{3,2}\\
Chenlin Ming\textsuperscript{3}, Keqi Zhu\textsuperscript{4,2}, Yizhi Chen\textsuperscript{5,2}, Shengcheng Fu\textsuperscript{5,2}, Zihao Zhou\textsuperscript{3,2}, Xinyi Peng\textsuperscript{5,2}\\
ZHANG Kaixuan\textsuperscript{6}, Boyang Xing\textsuperscript{7}, Jiawei Bao\textsuperscript{7}, Yang Yang\textsuperscript{8}, Yue Gao\textsuperscript{3,2,*}}
\newcommand{\PaperAffiliation}{%
\textsuperscript{1}Fudan University \quad \textsuperscript{2}Shanghai Innovation Institute \quad \textsuperscript{3}Shanghai Jiaotong University\\
\textsuperscript{4}Zhejiang University \quad \textsuperscript{5}Tongji University \quad \textsuperscript{6}The Chinese University of Hong Kong\\
\textsuperscript{7}Humanoid Robot (Shanghai) Co., Ltd. \quad \textsuperscript{8}Beihang University}
\newcommand{\PaperContact}{yuegao@sjtu.edu.cn}
\renewcommand{\LabName}{RL\textsuperscript{2} Lab}
\renewcommand{\LabLogoPath}{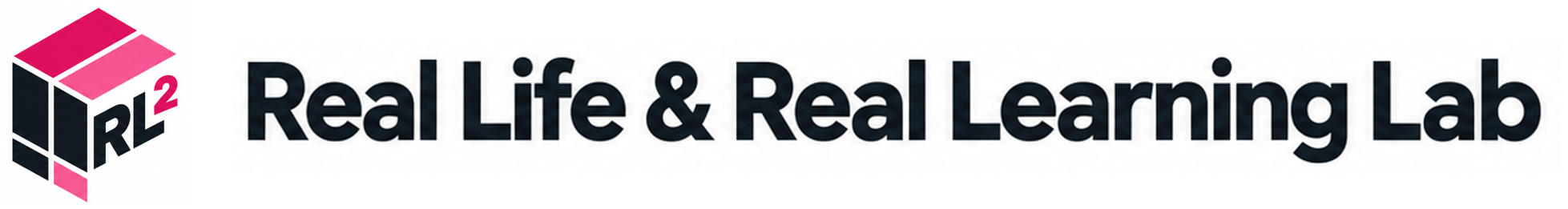}

\usepackage{amsmath,amsfonts,bm}

\def\eqref#1{equation~\ref{#1}}

\def\1{\bm{1}}

\DeclareMathAlphabet{\mathsfit}{\encodingdefault}{\sfdefault}{m}{sl}
\SetMathAlphabet{\mathsfit}{bold}{\encodingdefault}{\sfdefault}{bx}{n}

\newcommand{\papertablestyle}{\fontsize{8}{9.5}\selectfont\setlength{\tabcolsep}{2pt}\setlength{\aboverulesep}{1.2pt}\setlength{\belowrulesep}{1.2pt}\renewcommand{\arraystretch}{1}}

\newcommand{\PaperEndMatterSpacing}{\par}
\newcommand{\PaperTeaserWidth}{0.9\linewidth}
\hypersetup{pdftitle={\PaperTitle},pdfauthor={\PaperAuthorNames}}
\begin{document}
\title{\bfseries\PaperTitle}
\ifx\PaperAuthors\empty
\author{}
\else
\author{%
  \begin{tabular}{@{}>{\fontsize{10}{12}\selectfont}c@{}}\PaperAuthors\end{tabular}\\[4pt]
  \begin{tabular}{@{}>{\fontsize{9}{11}\selectfont}c@{}}\PaperAffiliation\end{tabular}\\[4pt]
  \fontsize{9}{11}\selectfont\textsuperscript{\textdagger}These authors contributed equally.\quad
  \textsuperscript{*}Corresponding author:
  \href{mailto:\PaperContact}{\texttt{\PaperContact}}}
\fi
\date{}
\maketitle
\thispagestyle{labfirst}
\vspace{-1cm}
\begin{figure}[!hb]
  \centering
  \IfFileExists{figures/teaser.pdf}{%
    \includegraphics[width=\PaperTeaserWidth,pagebox=cropbox]{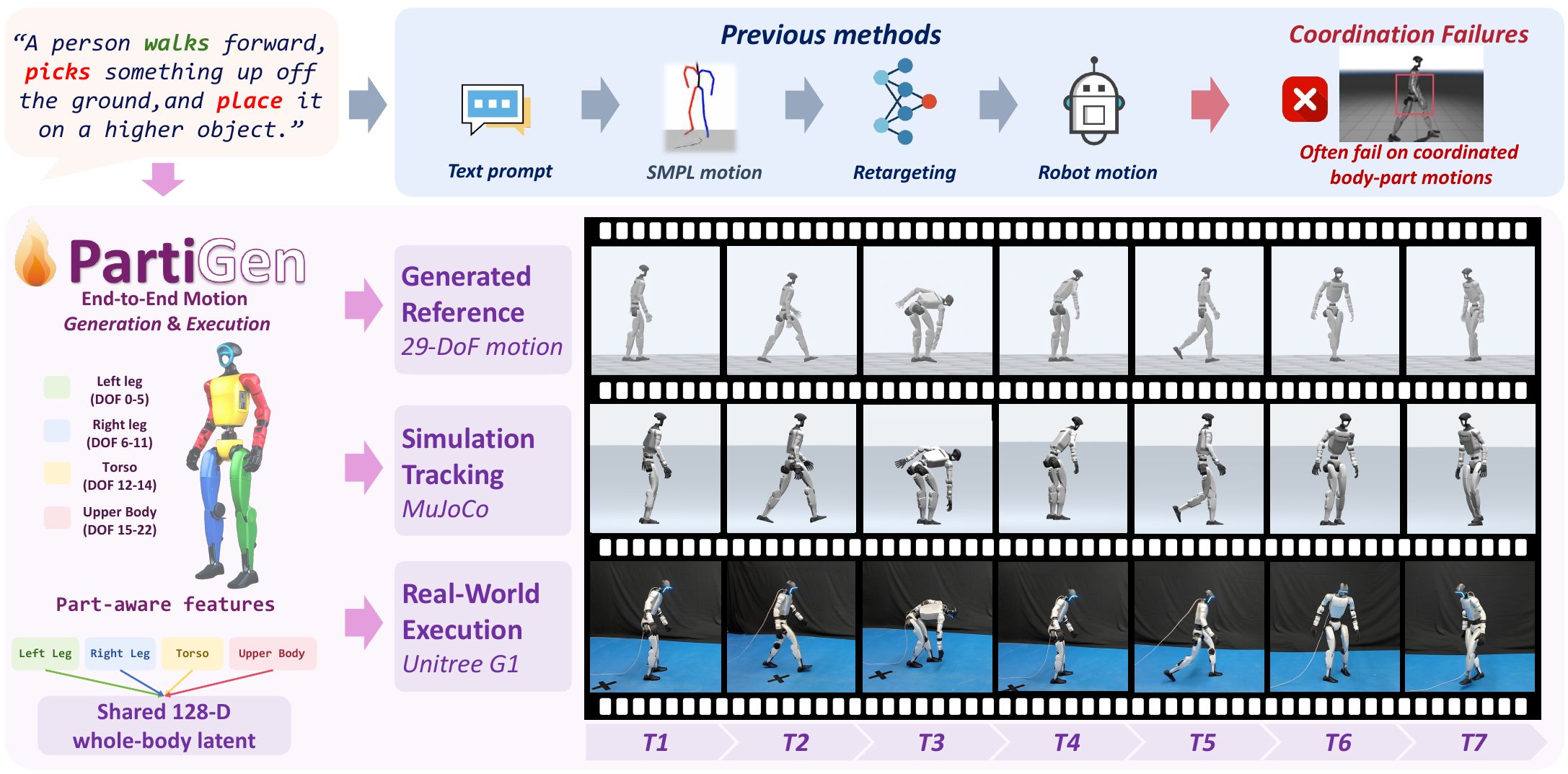}%
  }{%
    \fbox{\parbox[c][3.0cm][c]{0.96\linewidth}{\centering
      \textbf{Teaser artwork pending}\\[0.6em]
      Language instruction $\rightarrow$ Generated reference $\rightarrow$ MuJoCo tracking $\rightarrow$ Real-robot execution\\[0.6em]
      \small Reserved for \texttt{figures/teaser.pdf}}}%
  }
  \caption{\textbf{The PartiGen system: from language to motion and execution.} Our motion generator produces references; the execution component uses a pretrained, frozen GRIT controller~\citep{mrzuang2026grit} for tracking in MuJoCo and on the Unitree G1. Hardware examples are selected after simulation screening.}
  \label{fig:teaser}
\end{figure}

\begin{abstract}
Translating language into humanoid motion requires coordinating body parts, determining motion duration, and producing references suitable for robot execution. We introduce \textbf{PartiGen}, a part-aware latent diffusion framework that incorporates anatomical structure into a shared whole-body motion representation. Part-specific attention processes anatomically masked frame features, and its outputs are fused to combine local limb information with whole-body context. Conditioned on text and motion history, the model generates motion primitives autoregressively. Forward-kinematics supervision encourages kinematic consistency, while a learned end-of-sequence predictor selects a frame-level endpoint after generation. On HumanML3D retargeted to the Unitree G1, PartiGen achieves a segment-level FID of 3.165, a 16.8\% reduction over TextOp, and improves R@1 by 1.4 percentage points. In the duration study with 16-frame primitives, endpoint prediction reduces mean absolute duration error from 12.471 to 3.080 frames compared with a fixed-length cutoff. We further evaluate robot execution using a pretrained, frozen GRIT tracker: 409 of 500 generated references (81.8\%) pass the posture-based stability screen in MuJoCo. Following simulation screening and manual selection, 26 of 27 attempted motions are judged successful on the real G1. \textbf{Project page:} \url{https://andyhuangzz01.github.io/partigen-paper/}.

\end{abstract}

\section{Introduction}
\label{sec:intro}

Natural language offers an intuitive interface for specifying whole-body humanoid behaviors. Text-to-motion generation~\citep{guo2022generating,tevet2022human,guo2024momask} supports their synthesis, but visually plausible motion may violate robot joint limits, dynamics, actuator capabilities, or contact constraints. We study a system in which a language-conditioned generator specifies a whole-body reference and its temporal extent, while a tracking controller uses robot-state feedback to produce joint-level control targets. The challenge is to generate references that express the requested behavior while accounting for the robot's motion constraints and the task's temporal organization.

An instruction such as ``walk while waving'' illustrates the challenge: the arm gesture and locomotion must coexist within a single robot trajectory, with consistent root motion, joint configurations, and timing. A reference generator therefore needs to preserve the distinct roles of body parts while representing their relationships within a shared whole-body context. Part-aware modeling~\citep{zhong2023attt2m} and physics-guided generation~\citep{yuan2023physdiff} provide relevant foundations for addressing these requirements. We investigate how an anatomical prior can guide local feature processing within a shared robot-motion representation, together with kinematic supervision of the resulting trajectory. Its temporal extent must also reflect the instruction, with an endpoint that need not coincide with the boundary of a fixed-length motion primitive. These considerations connect body-part structure, kinematic consistency, and endpoint selection within the same robot reference-generation problem.

We propose \textbf{PartiGen}, integrating language-conditioned humanoid motion generation and execution. Part-aware attention builds a shared whole-body representation from anatomically masked inputs. Text- and history-conditioned latent diffusion generates motion primitives under forward-kinematics supervision, followed by frame-level endpoint selection on the complete candidate trajectory. A pretrained, frozen GRIT controller~\citep{mrzuang2026grit} tracks the references using robot-state feedback in MuJoCo and on the Unitree G1.

Our contributions are:
\begin{enumerate}
    \item \textbf{An anatomical input-grouping mechanism for robot motion representation learning.} Fixed masks select joint-angle channels before independent Q/K/V projections, while shared motion features and a single whole-body latent retain access to global context.
    \item \textbf{A language-conditioned motion generation framework with frame-level endpoint prediction.} Conditioned on text and motion history, the framework autoregressively generates motion primitives in the shared latent space under forward-kinematics supervision. A learned end-of-sequence predictor then selects an endpoint within the complete candidate trajectory.
    \item \textbf{Simulation and real-world validation.} Integrated with a frozen GRIT controller, PartiGen achieves an 81.8\% posture-screen pass rate over 500 simulated references. Following simulation screening and manual selection, 26 of 27 motions attempted on a physical Unitree G1 are judged successful by the authors.
\end{enumerate}

\section{Related Work}
\paragraph{Text-to-motion generation.}
Large-scale motion datasets~\citep{guo2022generating,AMASS:ICCV:2019,punnakkal2021babel} provide a foundation for data-driven generation. Common modeling paradigms include variational motion models~\citep{guo2020action2motion,petrovich2021action,petrovich22temos}, diffusion-based generation~\citep{tevet2022human,Zhang2022MotionDiffuse,zhang2023remodiffuse}, and discrete motion tokenization~\citep{guo2022tm2t,jiang2023motiongpt}. Motion representation learning~\citep{zhu2023motionbert} and controllable diffusion priors~\citep{karunratanakul2023guided,shafir2023human} offer complementary approaches to encoding and manipulating motion. DART and TextOp~\citep{Zhao:DartControl:2025,xie2025textop} further address autoregressive generation and interactive control. Building on autoregressive primitive generation, PartiGen incorporates anatomical input masks and EOS-based endpoint prediction.

\paragraph{Structure-aware modeling.}
Part-aware attention~\citep{zhong2023attt2m} and skeleton-aware latent diffusion~\citep{hong2025salad} incorporate anatomical structure into motion representations. Hierarchical models address fine-grained semantic control~\citep{jin2023act} and human-object interaction~\citep{Pi_2023_ICCV}, while cross-skeleton models accommodate varying skeletal structures~\citep{khani2025unimogen,xu2026necromancer,gat2025anytop}. Skeleton conditioning has also been explored for human image generation~\citep{ju2023humansd}. PartiGen places its anatomical prior at the attention input through fixed masks on raw frame features before part-specific Q/K/V projections. Each head receives its selected joint angles together with shared root/contact features and all joint-angle increments. Head outputs are fused into a shared stream, retaining a single motion latent rather than independent part latents.

\paragraph{Physical plausibility and robot execution.}
PhysDiff and InterDiff incorporate physical constraints into motion or interaction generation~\citep{yuan2023physdiff,xu2023interdiff}. Language-conditioned humanoid generation and control~\citep{jiang2024harmon,li2026from}, motion retargeting~\citep{joao2025gmr}, and whole-body teleoperation~\citep{ze2025twist} connect motion modeling to robot behavior. A2CF supports humanoid policy learning through state-dependent assistive forces that gradually diminish as proficiency improves~\citep{cao2025learning}. HiWET uses hierarchical reinforcement learning for world-frame end-effector tracking in long-horizon humanoid loco-manipulation~\citep{cao2026hiwet}. PartiGen combines FK-supervised motion generation with an execution component based on the pretrained, frozen GRIT controller~\citep{mrzuang2026grit}. Kinematic supervision regularizes references during generator training; execution evaluation measures their tracking under this adopted controller.

\paragraph{Temporal modeling.}
Sequence continuation and termination are distinct design choices. Related temporal generation frameworks include latent audio diffusion~\citep{liu2023audioldm}, neural-codec speech modeling~\citep{wang2023neural}, and autoregressive video diffusion~\citep{henschel2025streamingt2v}. For motion, PartiGen selects an endpoint at frame resolution. It generates a complete candidate sequence before applying an EOS probability threshold to select its endpoint; this selection does not stop generation online.

\section{Method}

\subsection{Problem Formulation}
Given a natural-language instruction $\mathbf{c}_{\text{text}}$ and a motion history $\mathbf{H}=\{\mathbf{x}_1,\ldots,\mathbf{x}_{n_H}\}$, our goal is to generate a future whole-body robot reference $\mathbf{F}=\{\mathbf{x}_{n_H+1},\ldots,\mathbf{x}_{n_H+n_F}\}$ and determine its length $n_F$. The reference should express the requested behavior through kinematically consistent motion with an appropriate duration. It is then converted into controller-compatible reference states and tracked by the system's controller using robot-state feedback.

Each frame is represented by a 57-dimensional vector that jointly describes the robot's root motion, foot-contact states, and joint configurations and their inter-frame changes:
\begin{equation}
\mathbf{x}_t = [\mathbf{o}_t,\mathbf{c}_t,\Delta\mathbf{r}^{\mathrm{local}}_t,h_t,\mathbf{q}_t,\Delta\mathbf{q}_t]\in\mathbb{R}^{57},
\end{equation}
where the five root orientation features are $\mathbf{o}_t=[\sin\phi_t,\cos\phi_t-1,\sin\theta_t,\cos\theta_t-1,\Delta\psi_t]$. Here, $\phi_t$ and $\theta_t$ denote root roll and pitch, and $\Delta\psi_t$ is the inter-frame yaw increment. The remaining components are left/right foot contacts $\mathbf{c}_t\in\{0,1\}^2$, local-frame root displacement $\Delta\mathbf{r}^{\mathrm{local}}_t\in\mathbb{R}^3$, root height $h_t\in\mathbb{R}$, joint angles $\mathbf{q}_t\in\mathbb{R}^{23}$, and inter-frame joint-angle increments $\Delta\mathbf{q}_t\in\mathbb{R}^{23}$. The increments are displacement features, not time-normalized velocities. With zero-based indexing, joint angles occupy channels 11--33 and their increments occupy channels 34--56.
\begin{figure*}[!t]
  \centering
  \includegraphics[width=0.94\textwidth]{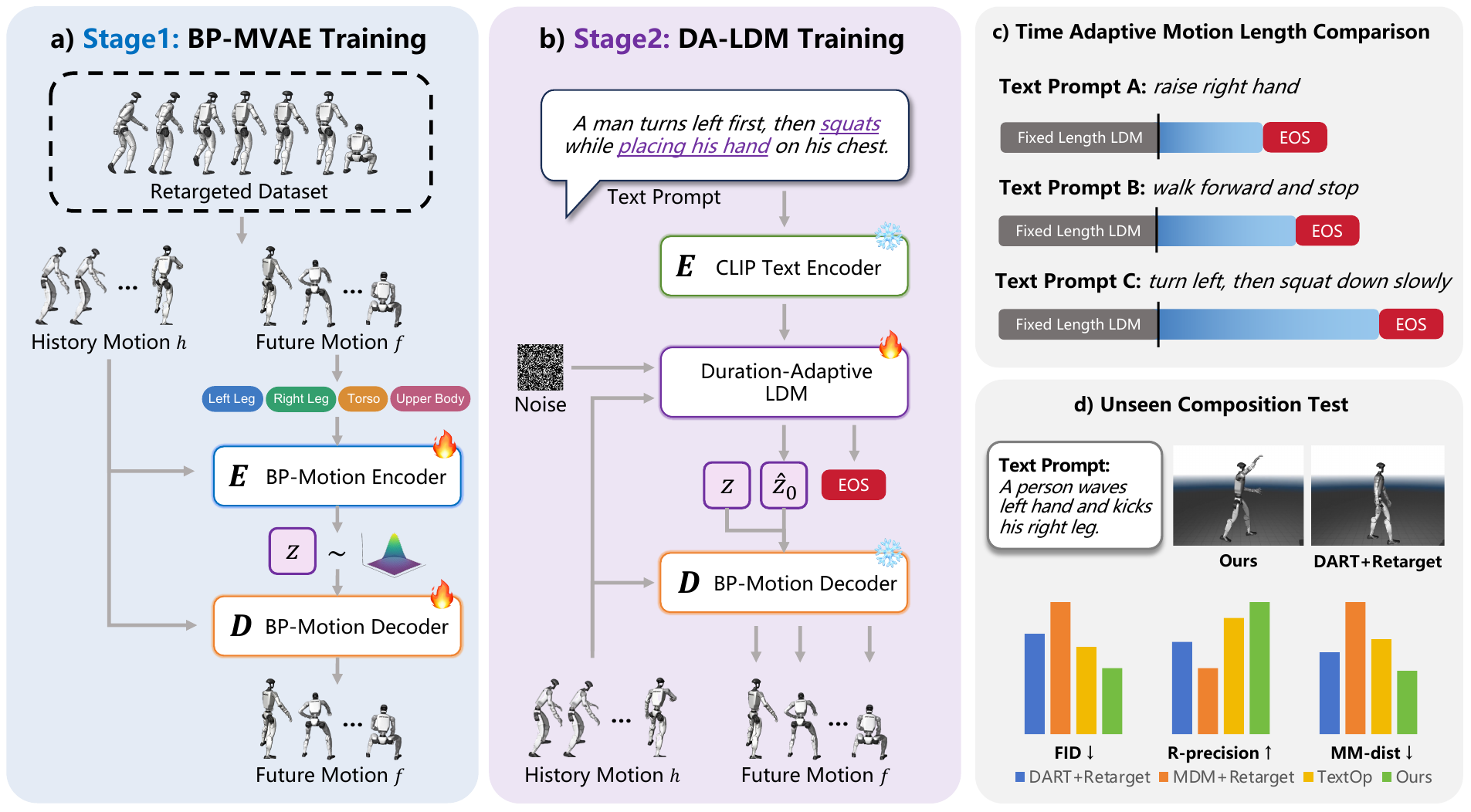}
  \caption{\textbf{Architecture and training of the PartiGen motion generator.} GRIT is the pretrained, frozen execution component described in Section~\ref{subsec:system_overview}.
  (a) \textbf{Stage 1: BP-MVAE Training.}
  Given history motion $h$ and future motion $f$, the BP-MVAE encodes motion into a shared latent vector $\mathbf{z}\in\mathbb{R}^{128}$. Part-aware features are fused into this vector for whole-body reconstruction.
  (b) \textbf{Stage 2: DA-LDM Training.}
  Building on latent diffusion~\citep{rombach2022high}, DA-LDM directly predicts the clean shared motion code ($x_0$ parameterization), conditioned on the noisy latent, CLIP text embeddings~\citep{radford2021learning}, history motion $h$, and diffusion timestep. The predicted clean latent is decoded to generate future motion $\hat{f}$.
  (c) \textbf{Endpoint Selection.} EOS selects an endpoint after complete candidate generation.
  (d) \textbf{Multi-Part Instructions.} Composition is evaluated as a diagnostic stress test, not an established zero-shot benchmark.}
  \label{fig:framework}
  \vspace{-0.3cm}
\end{figure*}

\subsection{System Overview: Generation and Execution}
\label{subsec:system_overview}
PartiGen generates language-conditioned whole-body reference trajectories for closed-loop tracking by a robot controller. {\bf Motion generation.} The generator is trained in two stages (Figure~\ref{fig:framework}): BP-MVAE first learns a shared 128-dimensional latent representation of whole-body motion, and DA-LDM then learns to generate motion codes in this space conditioned on text and motion history~\citep{rombach2022high}. At inference, successive codes are decoded into a complete candidate reference trajectory, after which EOS probabilities select its endpoint.

{\bf Motion execution.} For execution, the generated trajectories are mapped from 23 DoF at 30 Hz to 29 DoF at 50 Hz, with an interpolated transition into the first reference frame. A pretrained, frozen GRIT controller~\citep{mrzuang2026grit} tracks these references using robot-state feedback to produce joint-level control targets. Feedback remains within the tracking loop; generator training uses neither controller gradients nor a joint generation-control objective.

\subsection{Body-Part-Aware Motion VAE (BP-MVAE)}
A robot motion representation must capture local joint patterns together with their whole-body context. BP-MVAE uses part-aware attention to process anatomically masked frame features, then fuses the resulting features into a shared latent for whole-body reconstruction. The motion encoder injects anatomical structure before Q/K/V projection: fixed masks select each head's joint-angle channels from the raw 57-dimensional frame features, while root/contact features and all joint-angle increments remain shared. Each head applies independent Q/K/V projections and temporal attention, and the outputs are fused into the shared Transformer stream, combining part-specific inputs with whole-body context.

{\bf Shared 128-Dimensional Motion Latent.}
Each motion primitive is encoded as a single latent vector $\mathbf{z}\in\mathbb{R}^{128}$ that jointly represents root and joint motion. Anatomical structure guides feature processing, while all 128 latent coordinates are shared across the whole body. The decoder reconstructs the complete motion primitive from this code and motion history.

{\bf Kinematic Regularization.}
Joint-angle errors can propagate along the robot's kinematic chains, producing spatial deviations that feature-space losses do not directly measure. We therefore train the VAE with G1 forward-kinematics supervision and temporal smoothness regularization. Specifically, we enforce consistency in the kinematic state space via forward-kinematics-based supervision, together with temporal smoothness constraints:
\begin{equation}
\mathcal{L}_{\text{kin}} =
\rho\big( \Phi(\hat{\mathbf{x}}) - \Phi(\mathbf{x}) \big)+
\rho\big( \Delta^2 \hat{\mathbf{q}} \big),
\end{equation}
where $\Phi(\cdot)$ denotes FK-computed joint positions, rotations, and state variables, $\Delta^2$ represents the second-order temporal difference, and $\rho(\cdot)$ denotes the Huber loss. The overall VAE training objective is $\mathcal{L}_{\text{VAE}} = \mathcal{L}_{\text{rec}} + \beta \mathcal{L}_{\text{KL}} + \lambda_{\text{kin}} \mathcal{L}_{\text{kin}}$.

Each training run uses one NVIDIA RTX 4090 GPU with a batch size of 512 for both stages. We train both stages using the AdamW optimizer with a learning rate of $5.0 \times 10^{-5}$ and an annealing schedule. The gradient norm is clipped at 1.0 to ensure training stability. For the BP-MVAE, the model is trained for 200k steps, where we set the KL-divergence weight $\beta$ to $1.0 \times 10^{-4}$ and the kinematic loss weight $\lambda_{kin}$ to $0.8$. The DA-LDM is trained for 300k steps with an EOS loss weight $\lambda_{EOS}$ of $0.5$.
\subsection{Duration-Adaptive Latent Motion Diffusion (DA-LDM)}
\label{sec:da_ldm}
Conditioned on text and motion history, DA-LDM uses latent diffusion~\citep{rombach2022high} to generate successive motion codes in the shared space $\mathcal{Z}=\mathbb{R}^{128}$ learned by BP-MVAE. These codes are decoded into a candidate whole-body reference trajectory, whose endpoint is selected at frame resolution by EOS after generation. {\bf Shared-Latent Denoiser Architecture.}
We implement the denoiser $f_\theta$ using a Transformer backbone with direct clean-latent ($x_0$) prediction. At diffusion timestep $\tau$, its motion input is the noisy shared code $\mathbf{z}_{\tau}\in\mathbb{R}^{128}$. A common projection maps this code into the denoiser's hidden space:
\begin{equation}
\mathbf{u}_{\tau} = \mathbf{W}_{z}\mathbf{z}_{\tau} + \mathbf{b}_{z} + \mathbf{E}_{time}(\tau),
\end{equation}
where $\mathbf{W}_{z}\in\mathbb{R}^{512\times128}$ and $\mathbf{b}_{z}\in\mathbb{R}^{512}$ map the shared latent to the 512-dimensional hidden space, and $\mathbf{E}_{time}(\tau)$ is the diffusion timestep embedding. The denoiser operates on one shared motion code, without separate projections or latent branches for the root, arms, torso, or legs. {\bf Multi-Modal Conditioning.}
The denoiser is conditioned on motion history $\mathbf{H}$ via cross-attention, CLIP text features $\mathbf{c}_{\text{text}}$~\citep{radford2021learning}, and diffusion timestep $\tau$. For Classifier-Free Guidance (CFG)~\citep{ho2022classifier}, training sets the text condition to zero with probability $0.1$ while always retaining history. At inference, two forward evaluations yield clean-latent predictions with text present ($\hat{\mathbf{x}}_0^{\mathrm{text}}$) and text zeroed ($\hat{\mathbf{x}}_0^{\mathrm{zero}}$), using the same history. The sampler uses $\hat{\mathbf{x}}_0^{\mathrm{CFG}}=\hat{\mathbf{x}}_0^{\mathrm{zero}}+s(\hat{\mathbf{x}}_0^{\mathrm{text}}-\hat{\mathbf{x}}_0^{\mathrm{zero}})$ with $s=5.0$. {\bf Clean-Latent Training and Sampling.}
Let $\mathbf{x}_0=\mathbf{z}$ be the clean VAE latent. Training samples $\tau$ uniformly from $\{0,\ldots,4\}$ and forms $\mathbf{z}_{\tau}=\sqrt{\bar\alpha_{\tau}}\mathbf{x}_0+\sqrt{1-\bar\alpha_{\tau}}\boldsymbol{\epsilon}$, where $\boldsymbol{\epsilon}\sim\mathcal{N}(\mathbf{0},\mathbf{I})$. The network directly estimates $\hat{\mathbf{x}}_0=f_\theta(\mathbf{z}_{\tau},\tau,\mathbf{c}_{\text{text}},\mathbf{H})$. Its motion objective combines weighted clean-latent reconstruction, decoded future-motion reconstruction, and geometric supervision:
\begin{equation}
\mathcal{L}_{\text{motion}}=\mathbb{E}\!\left[w_{\tau}\left(\lambda_z\mathcal{L}_{\text{latent-rec}}+\lambda_f\mathcal{L}_{\text{future-rec}}+\lambda_g\mathcal{L}_{\text{geom}}\right)\right].
\end{equation}
Here $w_{\tau}=1$ for uniform timestep sampling and $\lambda_z=\lambda_f=\lambda_g=1$; $\mathcal{L}_{\text{geom}}$ is itself a weighted sum of geometry terms. Reconstruction and active geometry terms use mean Huber loss with $\delta=1$. Each training example uses one sampled noise level and one denoising prediction. DA-LDM is trained for 300k steps with ground-truth history (teacher forcing) and linear learning-rate annealing. The default sampler performs all five DDPM reverse updates with the guided clean-latent prediction and fixed posterior variance, followed by VAE decoding.
\subsection{Adaptive Motion Duration via EOS Prediction}
\label{sec:eos_prediction}
A requested behavior need not end at a fixed-length primitive boundary. Each primitive has a shared code $\mathbf{z}^{[k]}\in\mathbb{R}^{128}$ and generates $\ell\in\{8,16\}$ new frames; EOS therefore selects an endpoint at frame resolution after candidate generation. The motion head predicts the clean latent, while the EOS head is an MLP whose last hidden layer has width 128. EOS supervision uses positive-weighted binary cross-entropy,
\begin{equation}
\mathcal{L}_{\text{EOS}}=\operatorname{mean}_{\mathrm{valid}}
\left[-\alpha y\log\hat{s}-(1-y)\log(1-\hat{s})\right],\qquad \alpha=0.3,
\end{equation}
where each frame has a binary yes/no target $y$ and $\hat{s}$ is its predicted EOS probability; the mean is over valid frame targets. The positive-class coefficient is distinct from the overall EOS loss weight $\lambda_{\text{EOS}}=0.5$.
The overall training objective is $\mathcal{L}_{\text{total}} = \mathcal{L}_{\text{motion}} + \lambda_{\text{EOS}} \mathcal{L}_{\text{EOS}}$.

{\bf Post-Generation EOS Selection.}
All experiments generate a 320-frame candidate, matching the dataset maximum: 40 primitives for $\ell=8$ or 20 for $\ell=16$. The final motion reference retains frames up to and including the first frame whose EOS probability exceeds $\gamma=0.9$; if none does, all 320 frames are retained. Its length need not be a multiple of $\ell$. Selection occurs after complete candidate generation and does not stop generation early or save diffusion computation.
\section{Experiments}
We evaluate how well PartiGen generates robot references, including under multi-part instructions (Tables~\ref{tab:quantitative_comparison} and~\ref{tab:composition}), how its design choices affect reference quality and duration (Tables~\ref{tab:ablation} and~\ref{tab:eos}), and how a frozen controller executes the resulting motions (Section~\ref{subsec:real_robot} and Table~\ref{tab:tracking_preliminary}). Reference metrics are computed before tracking; ``Ours'' and ``Full'' denote the generator in these evaluations.
\subsection{Experimental Setup}

{\bf Motion Domains and Retargeting.}
TextOp provides a same-domain comparison: both it and PartiGen are trained on G1 motions and directly generate G1 references. DART~\citep{Zhao:DartControl:2025} and MDM~\citep{tevet2022human} provide generation-and-retargeting comparisons: their human-domain outputs are converted to SMPL as needed and retargeted to G1 using \textbf{General Motion Retargeting (GMR)}~\citep{joao2025gmr}, also used to construct HumanML3D-G1. All methods are evaluated as G1 references under the same metric protocol. DART+GMR and MDM+GMR comparisons therefore reflect differences in generation, training domain, and conversion path.

{\bf Text-Motion Evaluator.}
Due to the kinematic discrepancy between the G1 humanoid (23-DOF) and the SMPL human model, evaluators trained on human skeletons are not directly applicable. 
Following the HumanML3D evaluation protocol~\citep{guo2022generating}, we train a \textbf{G1-specific motion-text alignment model}. 
A GRU-based motion encoder and a pre-trained CLIP text encoder~\citep{radford2021learning} are used to map both modalities into a 512-dimensional shared latent space using contrastive learning. 
The generator and evaluator share a motion-disjoint split of approximately 25,000 training motions and 2,576 test motions; test motions are excluded from training. All methods use the same frozen evaluator checkpoint and identical text/motion preprocessing. The HumanML3D-G1 reference evaluation uses the complete test set for FID and Diversity. {\bf Implementation Details.}
BP-MVAE and DA-LDM are trained for 200k and 300k steps, respectively, each on a single RTX 4090 with batch size 512; evaluation uses batch size 128. Table~\ref{tab:quantitative_comparison} and the ablations use 8-frame primitives; the duration study uses 16-frame primitives for all variants.
\subsection{Quantitative Evaluation on HumanML3D-G1}

\begin{table}[t]
\caption{HumanML3D-G1 reference quality: segment (top) and transition (bottom).}
\label{tab:quantitative_comparison}
\centering
\papertablestyle

\begin{tabular*}{\linewidth}{@{\extracolsep{\fill}}lcccccc@{}}
\toprule
\textbf{Method} & R@1 $\uparrow$ & R@2 $\uparrow$ & R@3 $\uparrow$ & FID $\downarrow$ & Div $\rightarrow$ & mm-dist $\downarrow$ \\
\midrule
Dataset & $0.391 \!\pm\! 0.004$ & $0.475 \!\pm\! 0.005$ & $0.703 \!\pm\! 0.009$ & $0 \!\pm\! 0.000$ & $8.157 \!\pm\! 0.0035$ & $3.814 \!\pm\! 0.013$ \\
\midrule
DART+GMR& $0.230 \!\pm\! 0.007$ & $0.372 \!\pm\! 0.020$ & $0.486 \!\pm\! 0.018$ & $4.837 \!\pm\! 0.103$ & $\mathbf{8.012} \!\pm\! 0.069$ & $5.746 \!\pm\! 0.026$ \\
MDM+GMR & $0.186 \!\pm\! 0.004$ & $0.357 \!\pm\! 0.017$ & $0.473 \!\pm\! 0.018$ & $8.038 \!\pm\! 0.082$ & $7.883 \!\pm\! 0.046$ & $6.025 \!\pm\! 0.014$ \\
AttT2M+GMR & $0.157 \!\pm\! 0.003$ & $0.302 \!\pm\! 0.009$ & $0.438 \!\pm\! 0.013$ & $6.712 \!\pm\! 0.135$ & $\underline{8.323} \!\pm\! 0.071$ & $5.561 \!\pm\! 0.023$ \\
TextOp & $\underline{0.287} \!\pm\! 0.005$ & $\mathbf{0.398} \!\pm\! 0.009$ & $\underline{0.587} \!\pm\! 0.012$ & $\underline{3.802} \!\pm\! 0.145$ & $8.764 \!\pm\! 0.087$ & $\underline{4.464} \!\pm\! 0.059$ \\
Ours & $\mathbf{0.301} \!\pm\! 0.007$ & $\underline{0.386} \!\pm\! 0.014$ & $\mathbf{0.611} \!\pm\! 0.015$ & $\mathbf{3.165} \!\pm\! 0.068$ & $7.736 \!\pm\! 0.129$ & $\mathbf{4.308} \!\pm\! 0.086$ \\
\bottomrule
\end{tabular*}
\par\vspace{4pt}
\begin{tabular*}{\linewidth}{@{\extracolsep{\fill}}lccc@{}}
\toprule
\textbf{Method} & FID $\downarrow$ & Div $\rightarrow$ & PJ $\downarrow$ \\
\midrule
Dataset & $0 \!\pm\! 0.000$ & $6.223 \!\pm\! 0.024$ & $0.005 \!\pm\! 0.000$ \\
\midrule
DART+GMR & $\mathbf{3.925} \!\pm\! 0.114$ & $\underline{6.360} \!\pm\! 0.056$ & $\mathbf{0.009} \!\pm\! 0.000$ \\
MDM+GMR & $6.273 \!\pm\! 0.181$ & $5.991 \!\pm\! 0.042$ & $0.612 \!\pm\! 0.013$ \\
AttT2M+GMR & $6.633 \!\pm\! 0.102$ & $7.017 \!\pm\! 0.135$ & $0.357 \!\pm\! 0.004$ \\
TextOp & $4.273 \!\pm\! 0.125$ & $6.772 \!\pm\! 0.073$ & $0.023 \!\pm\! 0.001$ \\
Ours & $\underline{3.982} \!\pm\! 0.103$ & $\mathbf{6.266} \!\pm\! 0.052$ & $\underline{0.019} \!\pm\! 0.001$ \\
\bottomrule
\end{tabular*}
\vspace{-0.3cm}
\end{table}
We evaluate the overall generation quality on the HumanML3D dataset retargeted to the G1 humanoid robot.
{\bf Baselines.}
Reported baselines are \textbf{TextOp} (G1-domain), \textbf{DART+GMR}, \textbf{MDM+GMR}, and \textbf{AttT2M+GMR}~\citep{zhong2023attt2m}; the latter three use human-domain generation followed by GMR retargeting to G1.

{\bf Evaluation Protocol.}
Following the standard text-to-motion pipeline, we employ two sets of metrics:
(1) \textbf{Segment Evaluation}: R-Precision ($R@1/2/3$)~\citep{guo2022generating}, FID, Diversity (Div), and Multimodal Distance (mm-dist) to measure semantic alignment and distribution similarity.
(2) \textbf{Transition Evaluation}: FID, Diversity, and Position Jerk (PJ) to assess the smoothness and realism of motion transitions.
For Diversity ($\rightarrow$), closeness to the dataset value is the comparison criterion; a larger value is not necessarily better. Learned-method results reported as $\mathrm{mean}\pm\mathrm{SD}$ summarize three independent training-seed runs, evaluated with the same frozen evaluator; SD denotes standard deviation, not a confidence interval.

{\bf Results and Analysis.}
In Table~\ref{tab:quantitative_comparison}, PartiGen leads the evaluated generators in segment-level text-motion retrieval at \textbf{R@1 ($0.301$)} and \textbf{R@3 ($0.611$)}, and achieves the lowest \textbf{FID ($3.165$)} for distributional similarity. Relative to the same-domain TextOp baseline, FID decreases by 16.8\% (3.802 to 3.165) and R@1 increases by 1.4 percentage points (0.287 to 0.301). For transitions, PartiGen's diversity ($6.266$) is closest to the dataset value ($6.223$), while DART+GMR achieves lower FID and PJ.
\begin{figure*}[t]
  \centering
  \includegraphics[width=\linewidth,pagebox=cropbox]{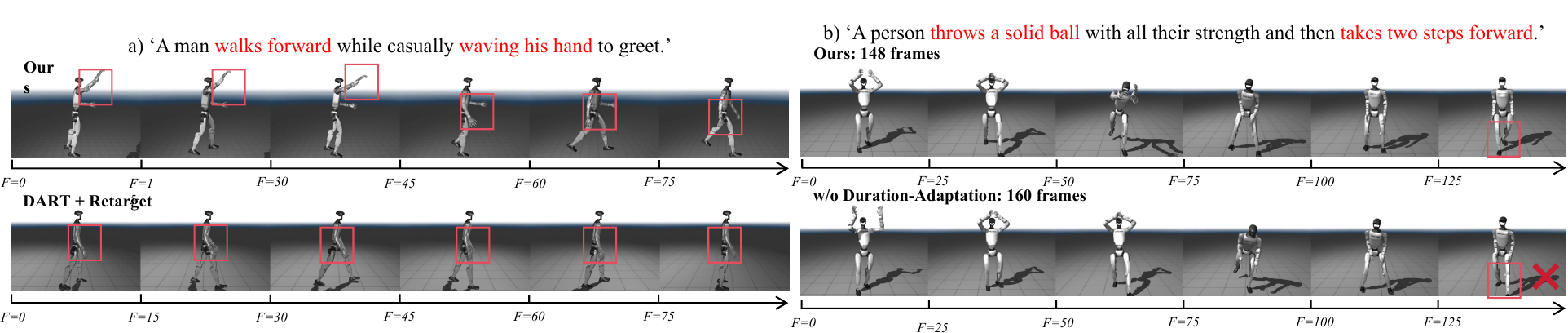}
  \caption{\textbf{Qualitative comparison.} (a) For ``walk while waving,'' the illustrated DART output shows a less pronounced arm gesture than PartiGen. (b) For ``throw then take two steps,'' PartiGen produces 148 frames after EOS-based endpoint selection, while the variant without duration adaptation uses a fixed length of 160 frames. These examples illustrate generated reference motions.}
  \label{fig:qualitative}
\end{figure*}

Figure~\ref{fig:qualitative} illustrates multi-part motion generation and EOS-based endpoint selection.

\subsection{Compositional Stress Test}
\begin{table}[t]
\caption{Generated-reference results on the compositional stress test (100 multi-part instructions).}
\label{tab:composition}
\centering
\papertablestyle

\begin{tabular*}{\linewidth}{@{\extracolsep{\fill}}lccccc@{}}
\toprule
\textbf{Method} & R@1 $\uparrow$ & R@2 $\uparrow$ & R@3 $\uparrow$ & FID $\downarrow$ & Div $\rightarrow$ \\
\midrule
Compositional Dataset & $0.331 \!\pm\! 0.006$ & $0.439 \!\pm\! 0.009$ & $0.634 \!\pm\! 0.013$ & $0.000 \!\pm\! 0.000$ & $6.102 \!\pm\! 0.127$ \\
\midrule
DART+GMR & $0.102 \!\pm\! 0.005$ & $0.148 \!\pm\! 0.012$ & $0.220 \!\pm\! 0.009$ & $9.340 \!\pm\! 0.125$ & $15.120 \!\pm\! 0.392$ \\
TextOp & $0.129 \!\pm\! 0.007$ & $0.143 \!\pm\! 0.018$ & $0.241 \!\pm\! 0.011$ & $9.951 \!\pm\! 0.174$ & $9.121 \!\pm\! 0.298$ \\
\textbf{Ours} & $\mathbf{0.211} \!\pm\! 0.008$ & $\mathbf{0.297} \!\pm\! 0.016$ & $\mathbf{0.413} \!\pm\! 0.021$ & $\mathbf{6.531} \!\pm\! 0.119$ & $\mathbf{8.693} \!\pm\! 0.167$ \\
\bottomrule
\end{tabular*}
\vspace{-0.4cm}
\end{table}
To probe whole-body robot reference generation when one instruction combines actions across body parts, we construct a diagnostic compositional stress test with 100 instructions.

{\bf Dataset Construction.}
Each sample describes coordinated actions across different body regions (e.g., ``walking while waving''). 
We synthesize reference motions by merging independent part-level trajectories: upper body, legs, and torso. 
We emphasize that this procedure yields only \emph{approximate} references: the merged trajectories are not captured from real coordinated performances and may contain inter-part inconsistencies, so the resulting numbers should be read as a diagnostic stress test rather than a curated benchmark.

{\bf Metrics.}
We report R-Precision, FID, and Diversity. 
FID and Diversity describe differences in the evaluator feature distribution relative to the synthesized references; they do not directly measure coordination or structural integrity. {\bf Discussion.}
In Table~\ref{tab:composition}, PartiGen obtains higher \textbf{R@1 ($0.211$)} than DART+GMR ($0.102$) and TextOp ($0.129$), alongside the lowest \textbf{FID ($6.531$)} and \textbf{diversity ($8.693$)} closest to the synthesized reference value ($6.102$). These gains indicate improved retrieval alignment and closer agreement with the synthesized reference distribution. Locomotion combined with large-amplitude arm gestures remains challenging across methods, and the approximate references limit conclusions about coordinated whole-body behavior. {\bf Experimental Takeaways.}
Together, the main evaluation and compositional stress test show gains in reference quality under the tested protocols. The following execution study assesses reference tracking, while the ablations examine the roles of anatomical grouping and kinematic supervision.

\subsection{System Evaluation: Closed-Loop Execution}
\label{subsec:real_robot}

To evaluate the execution of generated references, we compare tracking errors and posture-screen pass rates under the frozen GRIT controller for PartiGen, GT, DART+GMR, and Random-mask, with 500 references per group (Table~\ref{tab:tracking_preliminary}). GT and PartiGen use the same 29-DoF evaluation model, with the same six wrist joints set to zero.

\begingroup
\setlength{\intextsep}{6pt}
\begin{table}[htbp]
\centering
\papertablestyle
\caption{Frozen-GRIT tracking in MuJoCo. MPJPE: mean / median (mm), including valid unstable runs. Stability proxy: passes / requests.}
\label{tab:tracking_preliminary}

\begin{tabular*}{\linewidth}{@{\extracolsep{\fill}}lrrrrrcc@{}}
\toprule
Method & Requests & Valid & Excluded & \shortstack{Stability\\passes} & \shortstack{Stability\\rate ($\uparrow$)} & \shortstack{Global MPJPE\\(mm, $\downarrow$)} & \shortstack{Root-relative MPJPE\\(mm, $\downarrow$)} \\
\midrule
PartiGen (Ours) & 500 & 496 & 4 & 409 & 81.8\% & \textbf{252.402} / \textbf{108.871} & 57.995 / 39.765 \\
GT & 500 & 499 & 1 & 427 & 85.4\% & 399.78 / 183.22 & \textbf{56.24} / \textbf{37.94} \\
DART+GMR & 500 & 494 & 6 & 395 & 79.0\% & 288.35 / 241.03 & 58.32 / 40.07 \\
Random-mask & 500 & 494 & 6 & 401 & 80.2\% & 283.90 / 157.29 & 56.38 / 38.25 \\
\bottomrule
\end{tabular*}
\end{table}

\endgroup

Compared with Random-mask, Full has a higher observed posture-screen pass rate (81.8\% vs.\ 80.2\%) and lower Global MPJPE, while Random-mask has slightly lower root-relative MPJPE.

Across all valid motions, GT has a larger mean horizontal pelvis error at frame 0 after entry interpolation (142.84 vs.\ 16.44~mm). Correcting each trajectory by its fixed initial XYZ translation offset yields mean Global MPJPE of 308.89~mm for GT and 256.79~mm for PartiGen, narrowing the gap from 147.38 to 52.10~mm. Initial offsets thus explain a substantial portion of the raw gap in Table~\ref{tab:tracking_preliminary}; the remainder includes subsequent root-trajectory deviations and does not isolate motion quality.

The FK execution ablation yields a higher posture-screen pass rate for w/o FK (92.8\% vs.\ 81.8\%), while Full has lower tracking errors over valid runs. The screen does not check foot sliding, interpenetration, reference tracking, or semantic completion: a robot that remains standing without performing the requested motion can still pass. The pass-rate difference alone therefore cannot establish that Full has worse overall execution quality.

\begin{figure}[tbp]
  \centering
  \includegraphics[width=\linewidth,pagebox=cropbox]{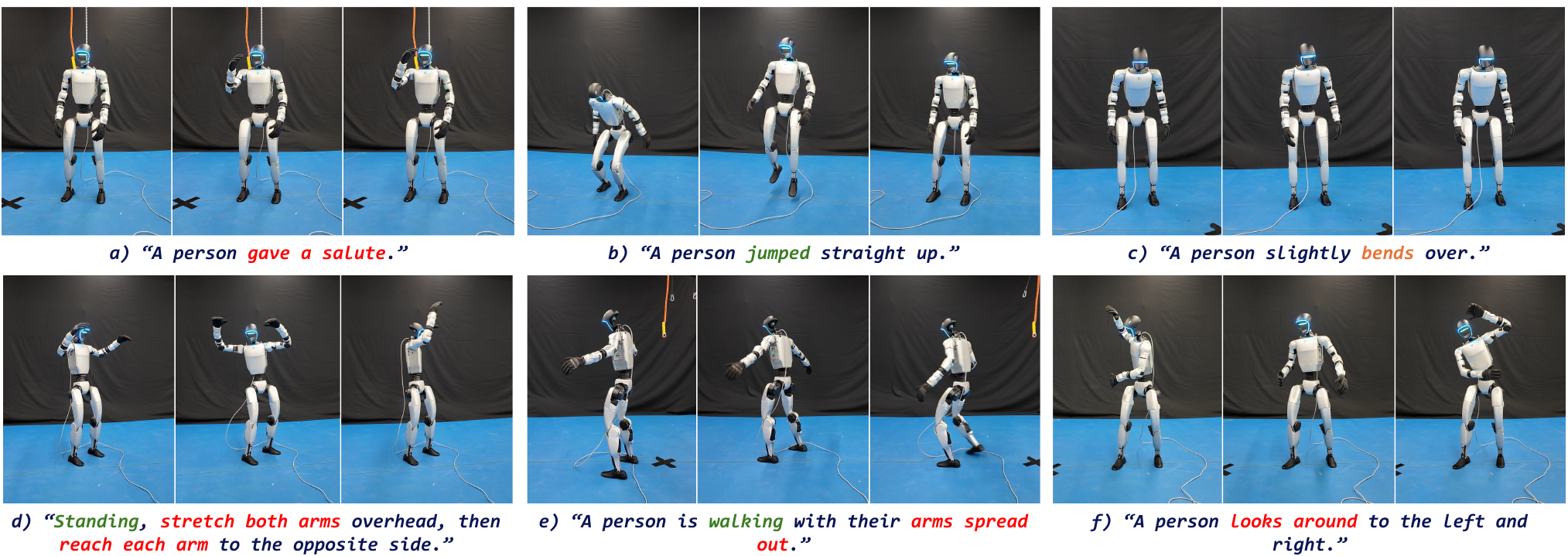}
  \caption{G1 executions selected after simulation screening and manual review.}
  \label{fig:hardware_examples}
  \vspace{-8pt}
\end{figure}

On the real G1, we evaluated the system's execution of generated references selected through simulation screening and manual review: the authors judged 26 of 27 attempted motions successful (96.3\%), while three additional selected motions were not attempted due to space constraints. Figure~\ref{fig:hardware_examples} shows selected hardware executions.

\subsection{Ablation Study}
We test whether anatomical grouping improves reference quality over random grouping with the same raw-feature projection structure. Data, training schedules, and hyperparameters are fixed. Section~\ref{sec:duration_study} evaluates EOS endpoint selection. We consider:
\textbf{Full (PartiGen)}: Complete generator with body-part-aware attention, duration adaptation, and FK-based geometric constraints. \textbf{w/o Part Attention}: Q/K/V are computed from the shared frame embedding using standard self-attention, without anatomical masks. This replaces the entire masked raw-feature attention module. \textbf{Random-mask}: Three random partitions (6/6/3/8), each fixed within its training run, retain the raw-feature projections.

{\bf Metrics.}
We report R-Precision, FID, Diversity, and position jerk (PJ) for reference quality and smoothness. Foot Contact Accuracy measures predicted versus ground-truth foot-contact agreement from FK-reconstructed trajectories. {\bf Analysis.}
With group sizes and raw-feature projections held fixed, Full achieves higher mean R@1 (0.301 vs.\ 0.281) and lower mean FID (3.165 vs.\ 3.533) than Random-mask (Table~\ref{tab:ablation}). This comparison supports the value of anatomical grouping in the tested configurations; w/o Part Attention evaluates replacement of the entire module.

\begin{table}[t]
\caption{Generator ablations on HumanML3D-G1. Random-mask summarizes three different partition/run combinations (mean $\pm$ sample SD). Contact Acc: foot-contact agreement.}
\label{tab:ablation}
\centering
\papertablestyle
\setlength{\tabcolsep}{1pt}
\resizebox{\linewidth}{!}{%
\begin{tabular}{@{}lccccccc@{}}
\toprule
Variant & R@1 $\uparrow$ & R@2 $\uparrow$ & R@3 $\uparrow$ & FID $\downarrow$ & Div $\rightarrow$ & PJ $\downarrow$ & \shortstack{Contact\\Acc $\uparrow$} \\
\midrule
\mbox{w/o Part Attention}
& $0.274 \!\pm\! 0.005$ 
& $0.356 \!\pm\! 0.009$ 
& $0.575 \!\pm\! 0.012$ 
& $3.811 \!\pm\! 0.094$ 
& $7.791 \!\pm\! 0.108$ 
& $0.035 \!\pm\! 0.001$
& $0.861 \!\pm\! 0.007$ \\

\mbox{Random-mask}
& $0.281 \!\pm\! 0.017$
& $0.362 \!\pm\! 0.009$
& $0.586 \!\pm\! 0.016$
& $3.533 \!\pm\! 0.218$
& $7.771 \!\pm\! 0.043$
& $0.028 \!\pm\! 0.009$
& $0.880 \!\pm\! 0.016$ \\

Full
& $\mathbf{0.301} \!\pm\! 0.007$ 
& $\mathbf{0.386} \!\pm\! 0.014$ 
& $\mathbf{0.611} \!\pm\! 0.015$ 
& $\mathbf{3.165} \!\pm\! 0.068$ 
& $7.736 \!\pm\! 0.129$ 
& $\mathbf{0.019} \!\pm\! 0.001$
& $\mathbf{0.914} \!\pm\! 0.005$ \\
\bottomrule
\end{tabular}%
}
\vspace{-0.4cm}
\end{table}



\subsection{Duration Modeling (EOS) Study}
\label{sec:duration_study}

We evaluate duration selection on HumanML3D-G1 using 16-frame primitives for all three variants, with other settings held fixed:
\textbf{Full (PartiGen)}: Our generator with EOS-based endpoint selection. \textbf{Fixed-Length}: EOS-based endpoint selection is replaced by a fixed cutoff at the dataset mean length after candidate generation. \textbf{Oracle Length}: Ground-truth motion lengths are provided during generation as a reference-length comparator, not a guaranteed upper bound on generation quality.

{\bf Metrics.}
We evaluate duration accuracy using \textbf{Duration Error}, defined as the absolute difference between predicted and ground-truth sequence lengths measured in frames.
We also report \textbf{R-Precision}, \textbf{FID}, and \textbf{Diversity} to assess motion quality and semantic alignment.

\begin{table}[!t]
\caption{Effect of duration modeling on generated references on HumanML3D-G1.}
\label{tab:eos}
\centering
\papertablestyle

\begin{tabular*}{\linewidth}{@{\extracolsep{\fill}}lcccc@{}}
\toprule
Variant & Duration Error $\downarrow$ & R@3 $\uparrow$ & FID $\downarrow$ & Div $\rightarrow$ \\
\midrule
Oracle Length & $\mathbf{0.000} \!\pm\! 0.000$ & $\mathbf{0.618} \!\pm\! 0.002$ & $\mathbf{2.980} \!\pm\! 0.015$ & 6.900 $\!\pm\!$ 0.045 \\
Fixed-Length  & 12.471 $\!\pm\!$ 3.596 & 0.584 $\!\pm\!$ 0.005 & 4.317 $\!\pm\!$ 0.021 & 6.760 $\!\pm\!$ 0.038 \\
Full (16-frame primitives) & 3.080 $\!\pm\!$ 0.385 & 0.599 $\!\pm\!$ 0.013 & 3.207 $\!\pm\!$ 0.033 & $\mathbf{7.836} \!\pm\! 0.048$ \\
\bottomrule
\end{tabular*}
\vspace{-0.4cm}
\end{table}

{\bf Analysis.}
Table~\ref{tab:eos} reports a lower duration error for Full than Fixed-Length (3.080 vs.\ 12.471 frames). Oracle Length has higher R@3 and lower FID than Full. This evaluation measures agreement with annotated lengths; a language instruction can admit multiple valid durations, so it does not establish an intrinsically optimal endpoint.

\section{Conclusion}
\label{sec:conclusion}
\textbf{PartiGen} integrates language-conditioned humanoid motion generation and closed-loop execution, using part-aware attention to learn a shared whole-body representation under kinematic supervision. Experiments support anatomical grouping in the tested configurations and show gains in segment-level reference quality and duration accuracy. A pretrained, frozen GRIT controller tracks the resulting references in MuJoCo and in selected G1 hardware executions.

\PaperEndMatterSpacing


\subsection*{AI use statement}

We used generative AI tools to polish the language and improve the clarity and readability of the manuscript. The authors take responsibility for all final content of this work.

\bibliographystyle{plainnat}
\bibliography{references}
\end{document}